\documentclass[letterpaper]{article}
\usepackage{iccc}

\usepackage{times}
\usepackage{helvet}
\usepackage{courier}

\usepackage{graphicx}
\usepackage{algorithm}
\usepackage{algpseudocode}
\usepackage{url}
\usepackage{ctable}
\usepackage[tight,footnotesize,nooneline]{subfigure}

\title{Automated Generation of Cross-Domain Analogies\\via Evolutionary Computation}
\author{At{\i}l{\i}m~G\"{u}ne\c{s}~Baydin$^{1,2}$, Ramon~L\'{o}pez~de~M\'{a}ntaras$^1$, Santiago Onta\~{n}\'{o}n$^3$\\
$^1$Artificial Intelligence Research Institute, IIIA\,-\,CSIC\\Campus Universitat Aut\`{o}noma de Barcelona, 08193 Bellaterra, Spain\\
$^2$Departament d'Enginyeria de la Informaci\'{o} i de les Comunicacions\\Universitat Aut\`{o}noma de Barcelona, 08193 Bellaterra, Spain\\
$^3$Department of Computer Science, Drexel University, 3141 Chestnut Street, Philadelphia, PA 19104, USA\\
gunesbaydin@iiia.csic.es, mantaras@iiia.csic.es, santi@cs.drexel.edu}
\begin{document}
\maketitle
\begin{abstract}
\begin{quote}
Analogy plays an important role in creativity, and is extensively used in science as well as art. In this paper we introduce a technique for the automated generation of cross-domain analogies based on a novel evolutionary algorithm (EA). Unlike existing work in computational analogy-making restricted to creating analogies between two given cases, our approach, for a given case, is capable of creating an analogy along with the novel analogous case itself. Our algorithm is based on the concept of ``memes'', which are units of culture, or knowledge, undergoing variation and selection under a fitness measure, and represents evolving pieces of knowledge as semantic networks. Using a fitness function based on Gentner's structure mapping theory of analogies, we demonstrate the feasibility of spontaneously generating semantic networks that are analogous to a given base network.
\end{quote}
\end{abstract}

\section{Introduction}

	In simplest terms, analogy is the transfer of information from a known subject (the \emph{analogue} or \emph{base}) onto another particular subject (the \emph{target}), on the basis of similarity. The cognitive process of analogy is considered at the heart of many defining aspects of human intellectual capacity, including problem solving, perception, memory, and creativity \cite{Holyoak1996}; and it has been even argued, by Hofstadter \shortcite{Hofstadter2001}, that analogy is ``the core of cognition''.

	Analogy-making ability is extensively linked with creative thought \cite{Hofstadter1995,Holyoak1996,Ward2001,Boden2004} and plays a fundamental role in discoveries and changes of knowledge in arts as well as science, with key examples such as Johannes Kepler's explanation of the laws of heliocentric planetary motion with an analogy to light radiating from the Sun\footnote{Kepler argued, as light can travel undetectably on its way between the source and destination, and yet illuminate the destination, so can motive force be undetectable on its way from the Sun to planet, yet affect planet's motion.} \cite{Gentner1997}; or Ernest Rutherford's analogy between the atom and the Solar System\footnote{The Rutherford--Bohr model of the atom considers electrons to circle the nucleus in orbits like planets around the Sun, with electrostatic forces providing attraction, rather than gravity.} \cite{Falkenhainer1989}. Boden \shortcite{Boden2004,Boden2009} classifies analogy as a form of \emph{combinational creativity}, noting that it works by producing unfamiliar combinations of familiar ideas.

	In this paper, we present a technique for the automated generation of cross-domain analogies using evolutionary computation. Existing research on computational analogy is virtually restricted to the discovery and assessment of analogies between a given pair of base case A and target case B \cite{French2002} (An exception is the Kilaza model by O'Donoghue \shortcite{ODonoghue2004}). On the other hand, given a base case A, the approach that we present here is capable of creating a novel analogous case B itself, along with the analogical mapping between A and B. This capability of open-ended creation of novel analogous cases is, to our knowledge, the first of its kind and makes our approach highly relevant from a computational creativity perspective. It replicates the psychological observation that an analogy is not always simply ``recognized'' between an original case and a retrieved analogous case, but the analogous case can sometimes be created together with the analogy \cite{Clement1988}.

	As the core of our approach, we introduce a novel evolutionary algorithm (EA) based on the concept of ``meme'' \cite{Dawkins1989}, where the individuals forming the population represent units of culture, or knowledge, that are undergoing variation, transmission, and selection. We represent individuals as simple semantic networks that are directed graphs of concepts and binary relations \cite{Sowa1991}. These go through variation by memetic versions of EA crossover and mutation, which we adapt to work on semantic networks, utilizing the commonsense knowledge bases of ConceptNet \cite{Havasi2007} and WordNet \cite{Fellbaum1998}. Defining a memetic fitness measure using analogical similarity from Gentner's psychological structure mapping theory \cite{Gentner1997}, we demonstrate the feasibility of generating semantic networks that are analogous to a given base network.

	In this introductory work, we focus on the evolution of analogies using a memetic fitness function promoting analogies. But it is of note that considering different possible fitness measures, the proposed representation and algorithm can serve as a generic tool for the generation of pieces of knowledge with any desired property that is a quantifiable function of the represented knowledge. Our algorithm can also act as a computational model for experimenting with memetic theories of knowledge, such as evolutionary epistemology and cultural selection theory.

	After a review of existing research in analogy, evolution, and creativity, the paper introduces details of our algorithm. We then present results and discussion of using the fitness function based on analogical similarity, and conclude with future work and potential applications in creativity.

\section{Background}
	\subsection{Analogy}

	Analogical reasoning has been actively studied from both cognitive and computational perspectives. The dominant school of research in the field, advanced by Gentner \cite{Falkenhainer1989,Gentner1997}, describes analogy as a structural matching, in which elements from a base domain are mapped to (or aligned with) those in a target domain via structural similarities of their relations. This approach named \emph{structure mapping theory}, with its computational implementation, the \emph{Structure Mapping Engine} (SME) \cite{Falkenhainer1989}, has been cited as the most influential work to date on the modeling of analogy-making \cite{French2002}. Alternative approaches in the field include the coherence based view developed by Holyoak and Thagard \cite{Thagard1990,Holyoak1996}, in which analogy is considered as a constraint satisfaction problem involving structure, semantic similarity, and purpose; and the view of Hofstadter \shortcite{Hofstadter1995} of analogy as a kind of high-level perception, where one situation is perceived as another one. Veale and Keane \shortcite{Veale1997} extend the work in analogical reasoning to the more specific case of metaphors, which describe the understanding of one kind of thing in terms of another. A highly related cognitive theory is the \emph{conceptual blending} idea developed by Fauconnier and Turner \shortcite{Fauconnier2002}, which involves connecting several existing concepts to create new meaning, operating below the level of consciousness as a fundamental mechanism of cognition. An implementation of this idea is given by Pereira \shortcite{Pereira2007} as a computational model of abstract thought, creativity, and language.

	According to whether the base and target cases belong to the same or different domains, there are two types of analogy: \emph{intra-domain}, confined to surface similarities within the same domain; and \emph{cross-domain}, using deep structural similarities between semantically distant information. While much of the research in artificial intelligence has been restricted to intra-domain analogies (e.g. case-based reasoning), studies in psychology have been more concerned with cross-domain analogical experiments \cite{Thagard1990}.

	\subsection{Evolutionary and Memetic Algorithms}

	Generalizing the mechanisms of the evolutionary process that has given rise to the diversity of life on earth, the approach of \emph{Universal Darwinism} uses a simple progression of variation, natural selection, and heredity to explain a wide variety of phenomena; and it extends the domain of this process to systems outside biology, including economics, psychology, physics, and even culture \cite{Dennett1995}. In terms of application, the metaheuristic optimization method of evolutionary algorithms (EA) provides an implementation of this idea, established as a solid technique with diverse problems in engineering as well as natural and social sciences \cite{CoelloCoello2007}.

	In an analogy with the unit of heredity in biological evolution, the gene, the concept of \emph{meme} was introduced by Dawkins \shortcite{Dawkins1989} as a unit of idea or information in cultural evolution, hosted, altered, and reproduced in individuals' minds, forming the basis of the field of memetics\footnote{Quoting Dawkins \cite{Dawkins1989}: \emph{``Examples of memes are tunes, ideas, catch-phrases, clothes fashions, ways of making pots or of building arches. Just as genes propagate themselves in the gene pool by leaping from body to body via sperms or eggs, so memes propagate themselves in the meme pool by leaping from brain to brain...''}}.

	Within evolutionary computation, the recently maturing field of memetic algorithms (MA) has experienced increasing interest as a method for solving many hard optimization problems \cite{Moscato2004}. The existing formulation of MA is essentially a hybrid approach, combining classical EA with local search, where the population-based global sampling of EA in each generation is followed by an individual learning step mimicking cultural evolution, performed by each candidate solution. For this reason, this approach has been often referred to under different names besides MA, such as ``hybrid EA'' or ``Lamarckian EA''. To date, MA has been successfully applied to a wide variety of problem domains such as NP-hard optimization problems, engineering, machine learning, and robotics.

	The potential of an evolutionary approach to creativity has been noted from cultural and practical viewpoints \cite{Gabora1997,Boden2009}. EA techniques have been shown to emulate creativity in engineering, such as genetic programming (GP) introduced by Koza \shortcite{Koza2003} as being capable of ``routinely producing inventive and creative results''\footnote{Striking examples of demonstrated GP creativity include replication of historically important discoveries in engineering, such as the reinvention of negative feedback circuits originally conceived by Harold Black in 1920s.}; as well as in visual art, design, and music \cite{Romero2008}. In psychology, there are studies providing support to an evolutionary view of creativity, such as the behavioral analysis by Simonton \shortcite{Simonton2003} inferring that scientific creativity constitutes a form of constrained stochastic behavior.

\section{The Algorithm}
	
	Our approach is based on a \emph{meme pool} comprising individuals represented as semantic networks, subject to variation and selection under a fitness measure. We position our algorithm as \emph{a novel memetic algorithm}, because (1) it is the units of culture, or information, that are undergoing variation, transmission, and selection, very close to the original sense of ``memetics'' as it was introduced by Dawkins; and (2) this is unlike the existing sense of the word in current MA as an hybridization of individual learning and EA. This algorithm is intended as a new tool focused exclusively on the memetic evolution of knowledge itself, which can find use in knowledge-based systems, reasoning, and creativity. 

	Our algorithm proceeds similar to a conventional EA cycle (Algorithm~\ref{AlgorithmNew}), with a relatively small set of parameters. We implement semantic networks as linked-list data structures of concept and relation objects. The descriptions of representation, fitness evaluation, variation, and selection steps are presented in the following sections. Parameters affecting each step of the algorithm are given in Table~\ref{TableParameters}.

	\begin{algorithm}
		\caption{Outline of the algorithm}
		\label{AlgorithmNew}
		\begin{algorithmic}[1]
		\Procedure{MemeticAlgorithm}{}
		\State $P(t=0) \gets$ \Call{Initialize}{$Pop_{size}, C_{max}, R_{min}, T$}
		\Repeat
		  \State $\phi(t) \gets$ \Call{EvaluateFitnesses}{$P(t)$}
		  \State $S(t) \gets$ \Call{Selection}{$P(t), \phi(t), S_{size}, S_{prob}$}
		  \State $V(t) \gets$ \Call{Variation}{$S(t), P_c, P_m, T$}
		  \State $P(t+1) \gets V(t)$
		  \State $t \gets t+1$
		\Until{stop criterion}
		\EndProcedure
		\end{algorithmic}
	\end{algorithm}

	\subsection{Representation}

	The algorithm is centered on the use of semantic networks \cite{Sowa1991} for encoding evolving memotypes. An important characteristic of a semantic network is whether it is definitional or assertional: in definitional networks the emphasis is on taxonomic relations (e.g. $IsA(bird, animal)$\footnote{Here we adopt the notation $IsA(bird, animal)$ to mean that the concepts $bird$ and $animal$ are connected by the directed relation $IsA$, i.e. ``bird is an animal.''}) describing a subsumption hierarchy that is true by definition; in assertional networks, relations describe instantiations that are contingently true (e.g. $AtLocation(human,$ $city)$). In this study we combine the two approaches for increased expressivity. As such, semantic networks provide a simple yet powerful means to represent the ``memes'' of Dawkins as data structures that are algorithmically manipulatable, allowing a procedural implementation of memetic evolution.

	In terms of representation, our approach is similar to several existing graph-based encodings of individuals in EA. The most notable is genetic programming (GP) \cite{Koza2003}, where candidate solutions are computer programs represented in a tree hierarchy. Montes and Wyatt \shortcite{Montes2004} present a detailed overview of graph-based EA techniques besides GP, which include parallel distributed genetic programming (PDGP), genetic network programming (GNP), evolutionary graph generation, and neural programming.

	Using a graph-based representation makes the design of variation operators specific to graphs necessary. In works such as GNP, this is facilitated by using a string-based encoding of node types and connectivity, permitting operators very close to their counterparts in conventional EA; and in PDGP, operations are simplified by making nodes occupy points in a fixed-size two-dimensional grid. What is common with GP related algorithms is that the output of each node in the graph can constitute an input to another node. In comparison, the range of connections that can form a semantic network of a given set of concepts is limited by commonsense knowledge, i.e. the relations have to make sense to be useful (e.g. $IsA(bird, animal)$ is meaningful while $Causes(bird, table)$ is not). To address this issue, we introduce new crossover and mutation operations for memetic variation, making use of commonsense reasoning \cite{Mueller2006} and adapted to work on semantic networks.

	\subsubsection{Commonsense Knowledge Bases}

	Commonsense reasoning refers to the type of reasoning involved in everyday thinking, based on \emph{commonsense knowledge} that an ordinary person is expected to know, or ``the knowledge of how the world works'' \cite{Mueller2006}. Knowledge bases such as the \emph{ConceptNet}\footnote{\url{http://conceptnet.media.mit.edu}} project of MIT Media Lab \cite{Havasi2007} and \emph{Cyc}\footnote{\url{http://www.cyc.com}} maintained by Cycorp company are set up to assemble and classify commonsense information. The lexical database \emph{WordNet}\footnote{\url{http://wordnet.princeton.edu}} maintained by the Cognitive Science Laboratory at Princeton University also has characteristics of a commonsense knowledge base, via synonym, hypernym\footnote{Y is a \emph{hypernym} of X if every X is a (kind of) Y ($IsA(dog, canine)$).}, and hyponym\footnote{Y is a \emph{hyponym} of X if every Y is a (kind of) X.} relations \cite{Fellbaum1998}.

	In our implementation we make use of ConceptNet version 4 and WordNet version 3 to process commonsense knowledge, where ConceptNet contributes around 560,000 definitional and assertional relations involving 320,000 concepts and WordNet contributes definitional relations involving around 117,000 synsets\footnote{A \emph{synset} is a set of synonyms that are interchangeable without changing the truth value of any propositions in which they are embedded.}. The hypernym and hyponym relations among noun synsets in WordNet provide a reliable collection of $IsA$ relations. In contrast, the variety of assertions in ConceptNet, contributed by volunteers across the world, makes it more prone to noise. We address this by ignoring all assertions with a reliability score (determined by contributors' voting) below a set minimum $R_{min}$ (Table~\ref{TableParameters}).

	\subsection{Initialization}

	At the start of each run of the algorithm, the population of size $Pop_{size}$ is initialized with individuals created by \emph{random semantic network generation} (Algorithm~\ref{AlgorithmNew}). This is achieved by starting from a network comprising only one concept randomly picked from commonsense knowledge bases and running a semantic network expansion algorithm that (1) randomly picks a concept in the given network (e.g. $human$); (2) compiles a list of relations---from commonsense knowledge bases---that the picked concept can be involved in (e.g. $\{CapableOf(human, think), Desires(human, eat), \cdots\}$) (3) appends to the network a relation randomly picked from this list, together with the other involved concept; and (4) repeats this until a given number of concepts has been appended or a set timeout $T$ has been reached (covering situations where there are not enough relations). Note that even if grown in a random manner, the resulting network itself is totally meaningful and consistent because it is a combination of rational information from commonsense knowledge bases.

	The initialization algorithm depends upon the parameters of $C_{max}$, the maximum number of initial concepts, and $R_{min}$, the minimum ConceptNet relation score (Table~\ref{TableParameters}).

	\subsection{Fitness Measure}

	Since the individuals in our approach represent knowledge, or memes, the fitness for evolutionary selection is defined as a function of the represented knowledge. For the automated generation of analogies through evolution, we introduce a memetic fitness based on analogical similarity with a given semantic network, utilizing the Structure Mapping Engine (SME) \cite{Falkenhainer1989,Gentner1997}. Taking the analogical matching score from SME as the fitness, our algorithm can evolve collections of information that are analogous to a given one.

  	In SME, an analogy is a one-to-one mapping from the base domain into the target domain, which correspond, in our fitness measure, to the semantic network supplied at the start and the individual networks whose fitnesses are evaluated by the function. The mapping is guided by the structure of relations between concepts in the two domains, ignoring the semantics of the concepts themselves; and is based on the systematicity principle, where connected knowledge is preferred over independent facts and is assigned a higher matching score. As an example, the Rutherford--Bohr atom and Solar System analogy \cite{Gentner1997} would involve a mapping from $sun$ and $planet$ in the first domain to $nucleus$ and $electron$ in the second domain. The labels and structure of relations in the two domains (e.g. $\{Attracts(sun, planet)$, $Orbits(planet, sun)$, $\cdots\}$ and $\{Attracts(nucleus$, $electron)$, $Orbits(electron$, $nucleus)$, $\cdots\}$) define and constrain the possible mappings between concepts that can be formed by SME.

  	We make use of our own implementation of SME based on the original description by Falkenhainer et al. \shortcite{Falkenhainer1989} and adapt it to the simple concept--relation structure of semantic networks, by mapping the predicate calculus constructs of \emph{entities} into \emph{concepts}, \emph{relations} to \emph{relations}, \emph{attributes} to $IsA$ \emph{relations}, and excluding \emph{functions}.

	\subsection{Selection}

	After assigning fitness values to all individuals in the current generation, these are replaced with offspring generated by variation operators on parents. The parents are probabilistically selected from the population according to their fitness, with reselection allowed. While individuals with a higher fitness have a better chance of being selected, even those with low fitness have a chance to produce offspring, however small. In our experiments we employ tournament selection \cite{CoelloCoello2007}, meaning that for each selection, a ``tournament'' is held among a few randomly chosen individuals, and the more fit individual of each successive pair is the winner according to a winning probability (Table~\ref{TableParameters}).

	In each cycle of algorithm, crossover is applied to parents selected from the population until $Pop_{size} \times P_c$ offspring are created (Table~\ref{TableParameters}). Mutation is applied to $Pop_{size} \times P_m$ selected individuals, supplying the remaining part of the next generation (i.e. $P_c + P_m = 1$). We also employ elitism, by replacing a randomly picked offspring in next generation with the individual with the current best fitness.

	\subsection{Variation Operators}

	In contrast with existing graph-based evolutionary approaches that we have mentioned, our representation does not permit arbitrary connections between different nodes and requires variation operators that should be based on information provided by commonsense knowledge bases. This means that any variation operation on the individuals should: (1) preserve the structure within boundaries set by commonsense knowledge; and (2) ensure that even vertices and edges randomly introduced into a semantic network connect to existing ones through meaningful and consistent relations\footnote{It should be noted that we depend on the meaningfulness and consistency (i.e. compatibility of relations with others involving the same concepts) of information in the commonsense knowledge bases, which should be ensured during their maintenance.}.

	Here we present commonsense crossover and mutation operators specific to semantic networks.

	\subsubsection{Commonsense Crossover}

	In classical EA, features representing individuals are commonly encoded as linear strings and the crossover operation simulating genetic recombination is simply a cutting and merging of this one dimensional object from two parents. In graph-based approaches such as GP, subgraphs can be freely exchanged between parent graphs \cite{Koza2003,Montes2004}. Here, as mentioned, the requirement that a semantic network has to make sense imposes significant constraints on the nature of recombination.

	We introduce two types of \emph{commonsense crossover} that are tried in sequence by the variation algorithm. The first type attempts a sub-graph interchange between two selected parents similar to common crossover in standard GP; and where this is not feasible due to the commonsense structure of relations forming the parents, the second type falls back to a combination of both parents into a new offspring.

\begin{figure*}[!t]
	    \centering
	    \subfigure[Parent 1]{\includegraphics[width=1.8in]{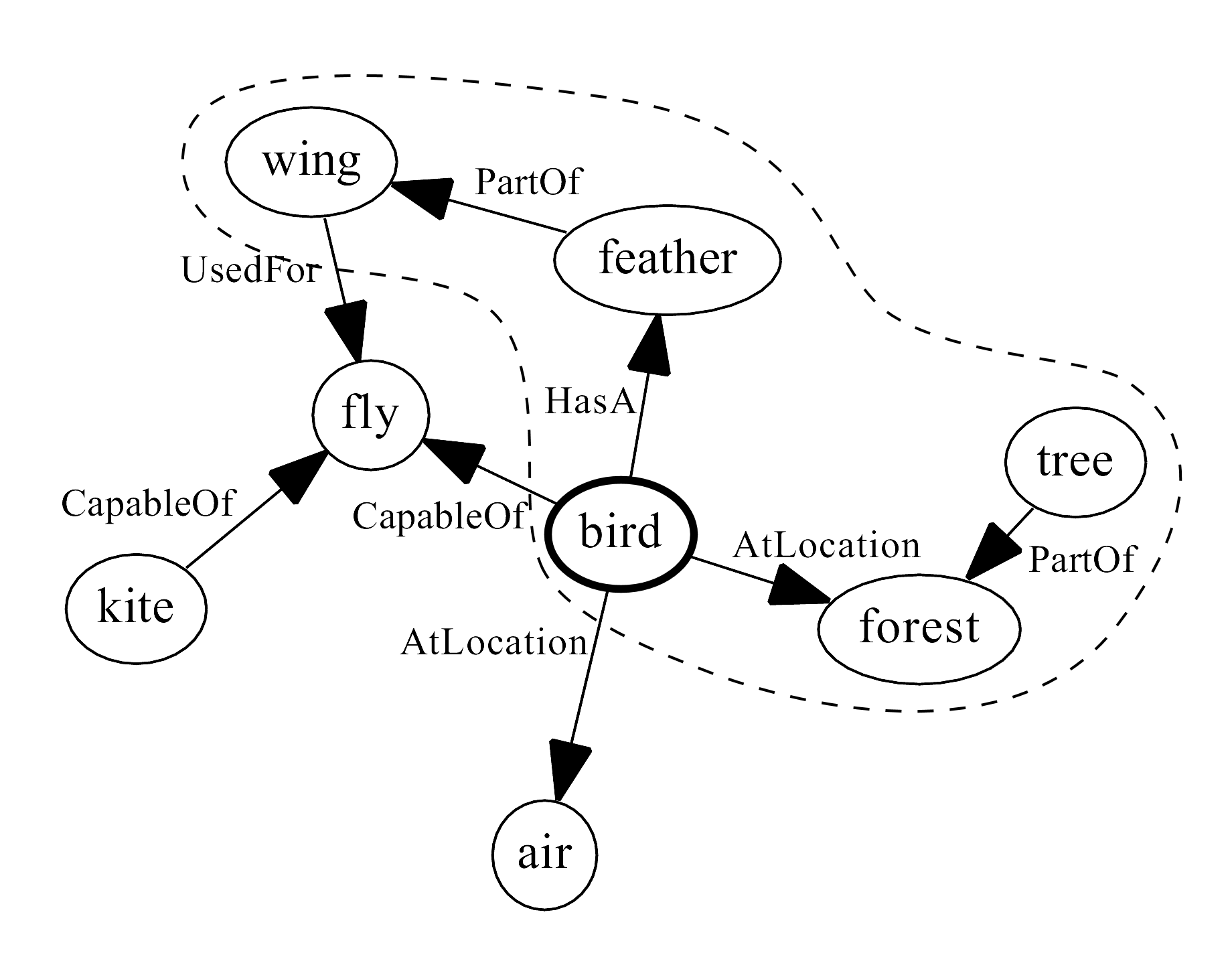}}
	    \hfil
	    \subfigure[Parent 2]{\includegraphics[width=1.8in]{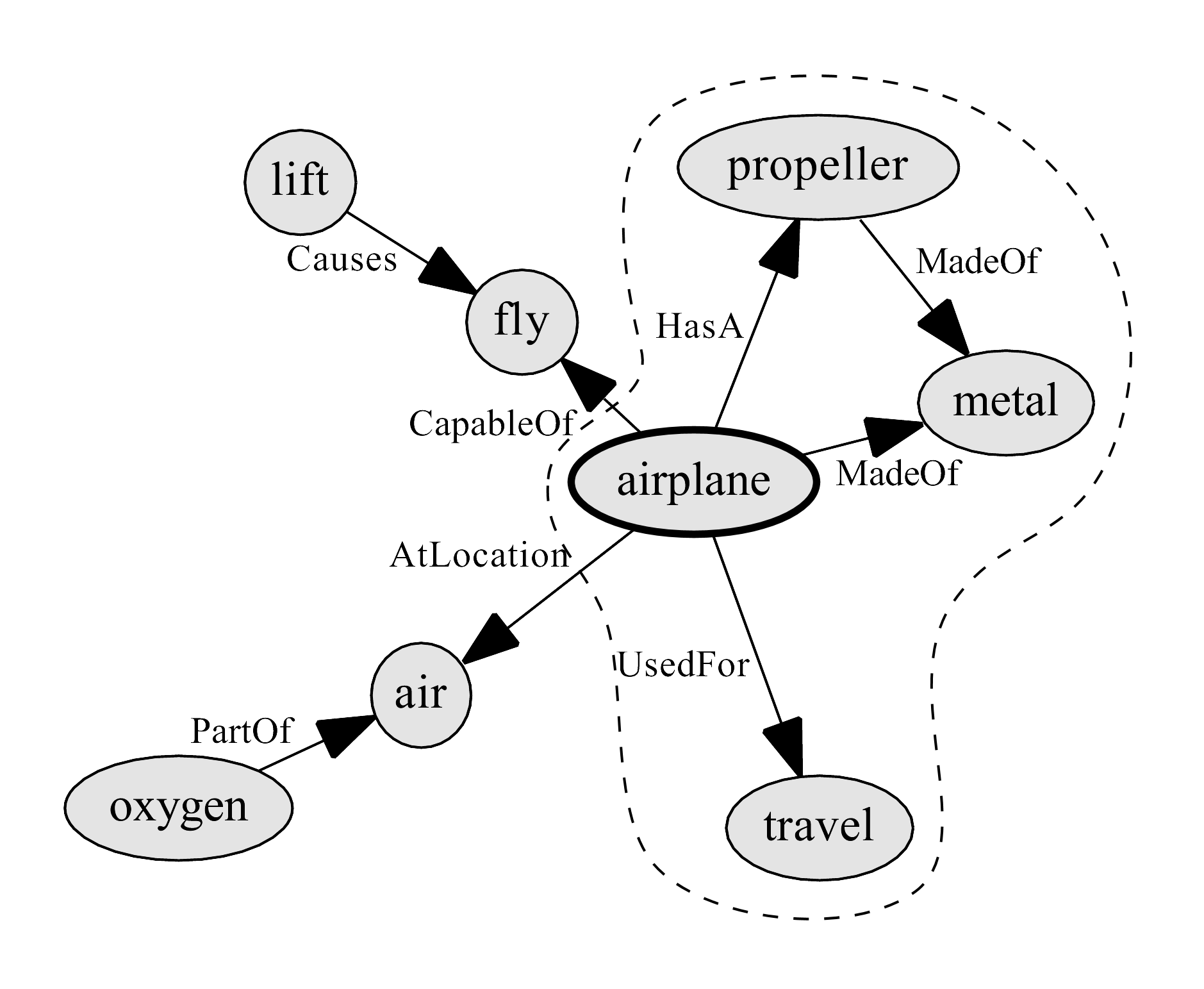}}\\
	    \subfigure[Offspring 1]{\includegraphics[width=1.7in]{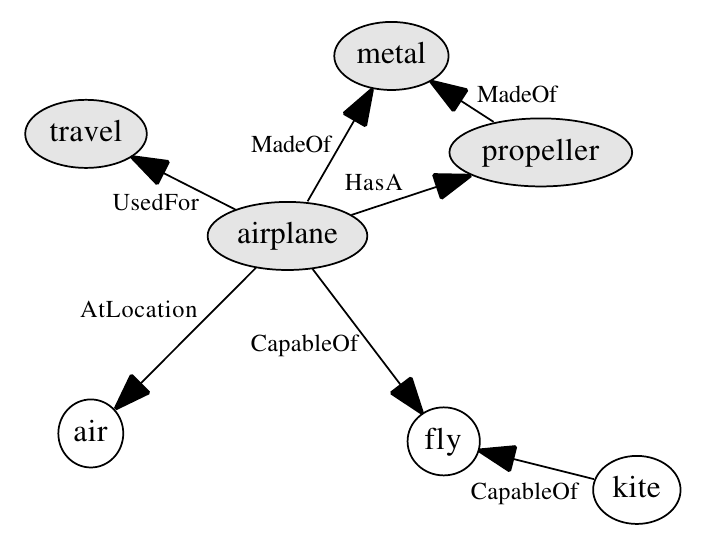}}
	    \hfil
	    \subfigure[Offspring 2]{\includegraphics[width=1.9in]{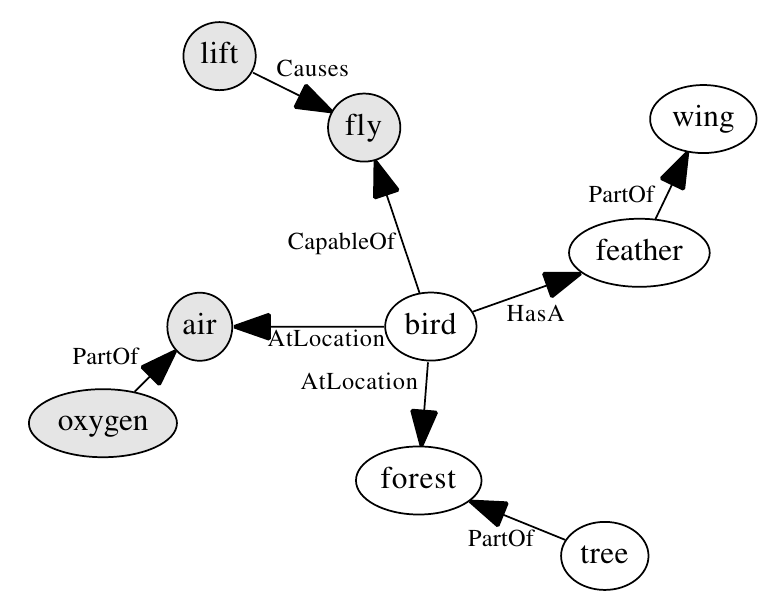}}
	    \caption{Commonsense crossover type I (subgraph crossover), centered on the concepts of $bird$ for parent 1 and $airplane$ for parent 2.}
	    \label{FigureCrossoverTypeI}
  	\end{figure*}

	\emph{Type I (subgraph crossover):} A pair of concepts, one from each parent, that are \emph{interchangeable}\footnote{We define two concepts from different semantic networks as \emph{interchangeable} if both can replace the other in all, or part, of the relations the other is involved in, queried from commonsense knowledge bases.} are selected as \emph{crossover concepts}, picked randomly out of all possible such pairs. For instance, in Figure~\ref{FigureCrossoverTypeI}, $bird$ and $airplane$ are interchangeable, since they can replace each other in the relations $CapableOf(\cdot, fly)$ and $AtLocation(\cdot, air)$. In each parent, a subgraph is formed, containing: (1) the crossover concept; (2) the set of all relations, and associated concepts, that are not common with the other crossover concept (In Figure~\ref{FigureCrossoverTypeI} (a), $HasA(bird, feather)$ and $AtLocation(bird, forest)$; and in (b) $HasA(airplane, propeller)$, $MadeOf(airplane, metal)$, and $UsedFor(airplane, travel)$); and (3) the set of all relations and concepts connected to these (In Figure~\ref{FigureCrossoverTypeI} (a) $PartOf(feather, wing)$ and $PartOf(tree, forest)$; and in (b) $MadeOf(propeller, metal)$), excluding the ones that are also one of those common with the other crossover concept (the concept $fly$ in Figure~\ref{FigureCrossoverTypeI} (a), because of the relation $CapableOf(\cdot, fly)$). This, in effect, forms a subgraph of information specific to the crossover concept, which is insertable into the other parent. Any relations between the subgraph and the rest of the network not going through the crossover concept are severed (e.g. $UsedFor(wing, fly)$ in Figure~\ref{FigureCrossoverTypeI} (a)). The two offspring are formed by exchanging these subgraphs between the parent networks (Figure~\ref{FigureCrossoverTypeI} (c) and (d)).

\begin{figure*}[!t]
	    \centering
	    \subfigure[Parent 1]{\includegraphics[width=1.45in]{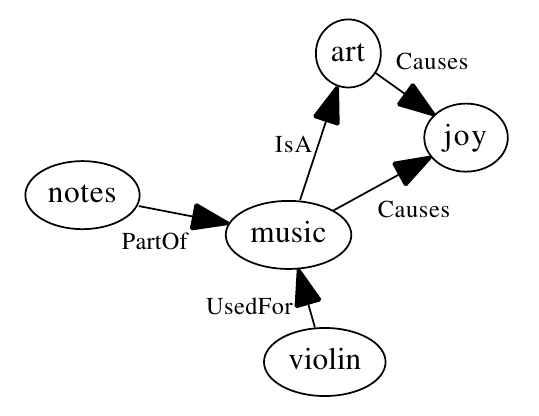}}
	    \hfil
	    \subfigure[Parent 2]{\includegraphics[width=1.4in]{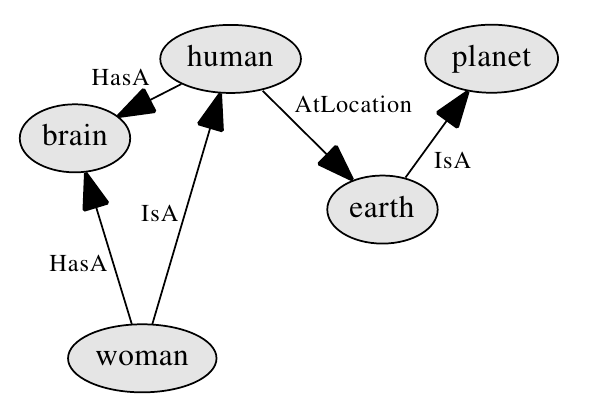}}
	    \hfil
	    \subfigure[Offspring]{\includegraphics[width=1.5in]{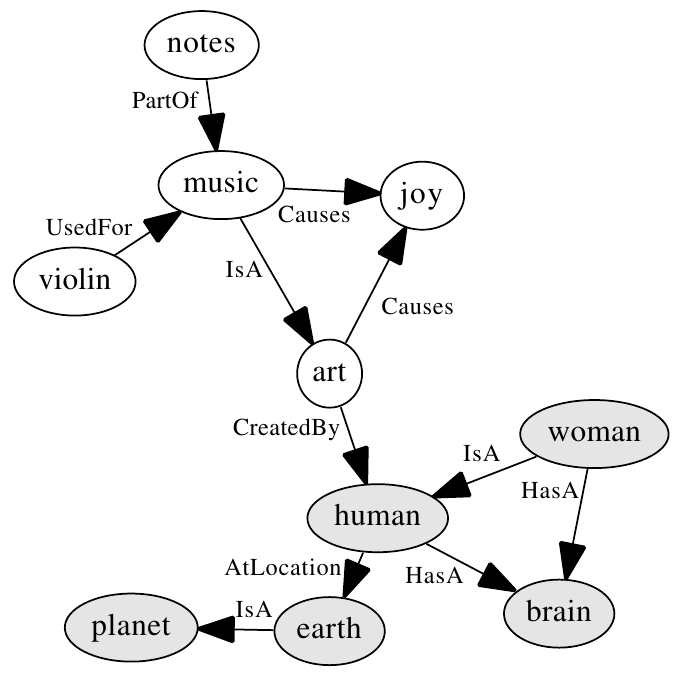}}
	    \caption{Commonsense crossover type II (graph merging crossover), merging by the relation $CreatedBy(art, human)$. If no concepts attachable through commonsense relations are encountered, the offspring is formed by merging the parent networks as two separate clusters within the same semantic network.}
	    \label{FigureCrossoverTypeII}
	\end{figure*}

	\emph{Type II (graph merging crossover):} A concept from each parent that is \emph{attachable}\footnote{We define a distinct concept as \emph{attachable} to a semantic network if at least one commonsense relation connecting the concept to any of the concepts in the network can be discovered from commonsense knowledge bases.} to the other parent is selected as a \emph{crossover concept}. The two parents are merged into an offspring by attaching a concept in one parent to another concept in the other parent, picked randomly out of all possible attachments ($CreatedBy(art, human)$ in Figure~\ref{FigureCrossoverTypeII}. Another possibility is $Desires(human, joy)$.). The second offspring is formed randomly the same way. In the case that no attachable concepts are found, the parents are merged as two separate clusters within the same semantic network. 

	\subsubsection{Commonsense Mutation}

	We introduce several types of \emph{commonsense mutation} operators that modify a parent by means of information from commonsense knowledge bases. For each mutation to be performed, the type is picked at random with uniform probability. If the selected type of mutation is not feasible due to the commonsense structure of the parent, another type is again picked. In the case that a set timeout of $T$ trials has been reached without any operation, the parent is returned as it is.

\begin{figure*}[!t]
		\centering
	    \subfigure[Mutation type I (before)]{\parbox[t]{1.6in}{\centering\includegraphics[width=1.4in]{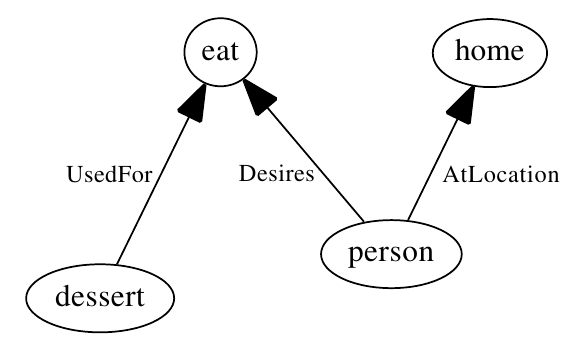}}}
	    \hfil
	    \subfigure[Mutation type I (after)]{\parbox[t]{1.6in}{\centering\includegraphics[width=1.4in]{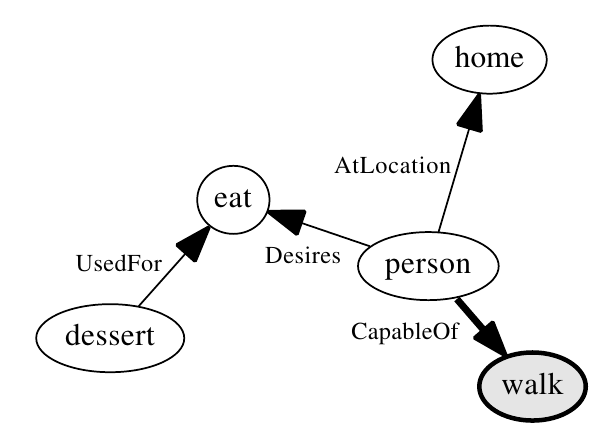}}}
	    \hfil
	    \subfigure[Mutation type IIa (before)]{\parbox[t]{1.6in}{\centering\includegraphics[width=1.1in]{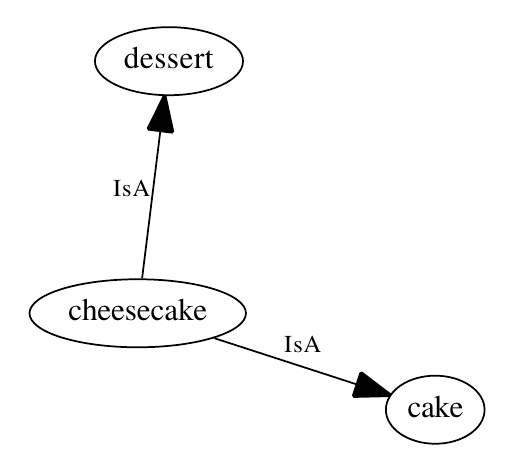}}}
	    \hfil
	    \subfigure[Mutation type IIa (after)]{\parbox[t]{1.6in}{\centering\includegraphics[width=1.1in]{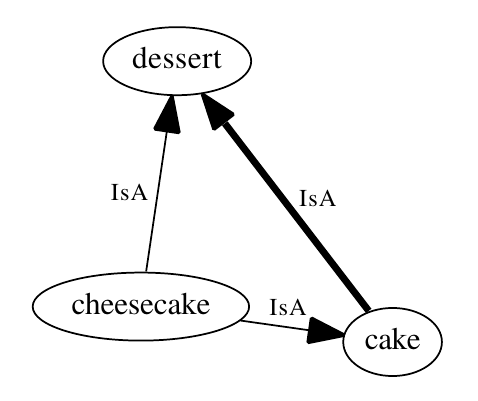}}}
	    \hfil
	    \subfigure[Mutation type IIb (before)]{\parbox[t]{1.6in}{\centering\includegraphics[width=1.0in]{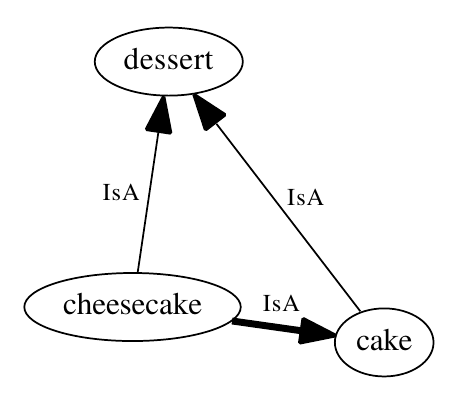}}}
	    \hfil
	    \subfigure[Mutation type IIb (after)]{\parbox[t]{1.6in}{\centering\includegraphics[width=1.0in]{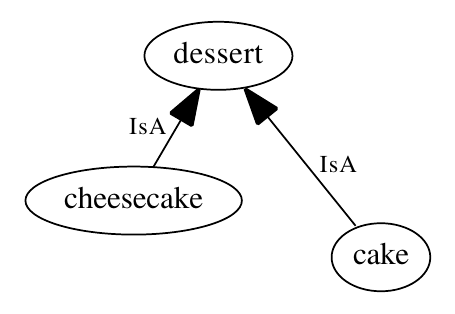}}}
	    \hfil
	    \subfigure[Mutation type IIIa (before)]{\parbox[t]{1.6in}{\centering\includegraphics[width=1.4in]{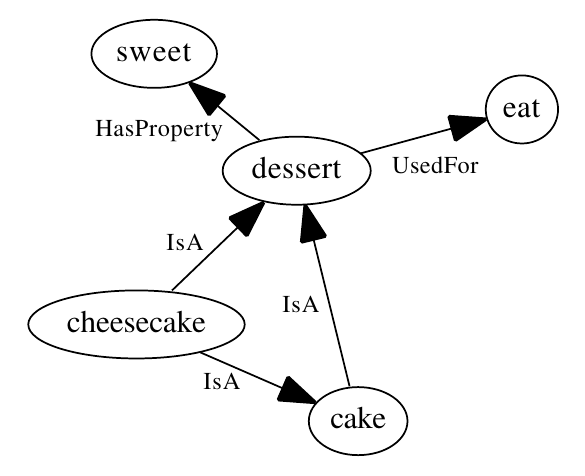}}}
	    \hfil
	    \subfigure[Mutation type IIIa (after)]{\parbox[t]{1.6in}{\centering\includegraphics[width=1.4in]{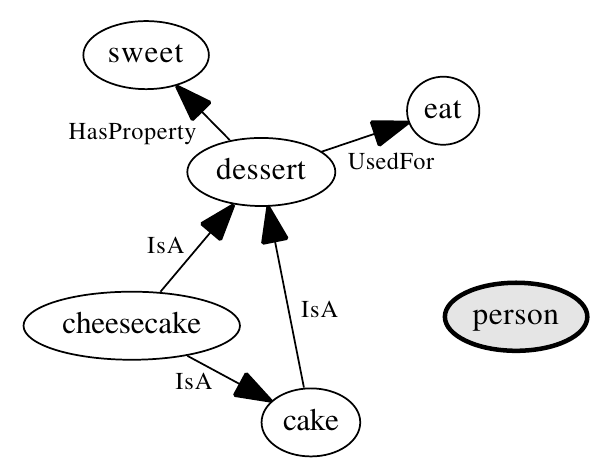}}}
	    \hfil
	    \subfigure[Mutation type IIIb (before)]{\parbox[t]{1.6in}{\centering\includegraphics[width=1.3in]{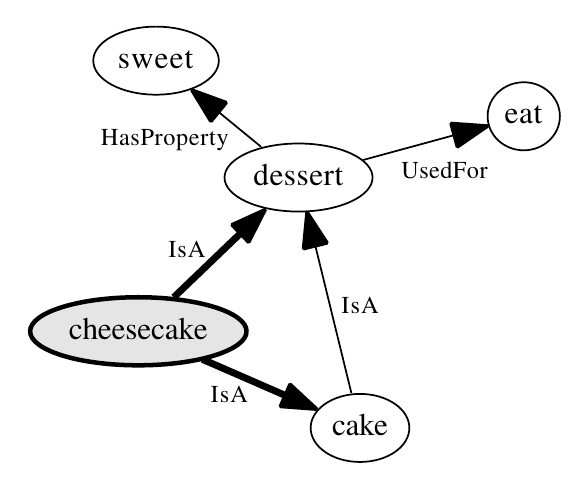}}}
	    \hfil
	    \subfigure[Mutation type IIIb (after)]{\parbox[t]{1.6in}{\centering\includegraphics[width=1.3in]{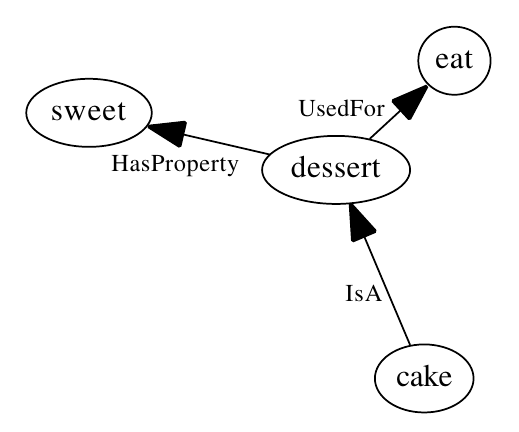}}}
	    \hfil
	    \subfigure[Mutation type IV (before)]{\parbox[t]{1.6in}{\centering\includegraphics[width=1.3in]{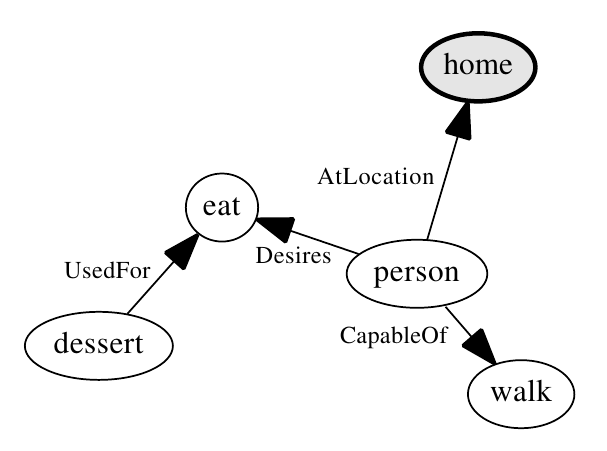}}}
	    \hfil
	    \subfigure[Mutation type IV (after)]{\parbox[t]{1.6in}{\centering\includegraphics[width=1.3in]{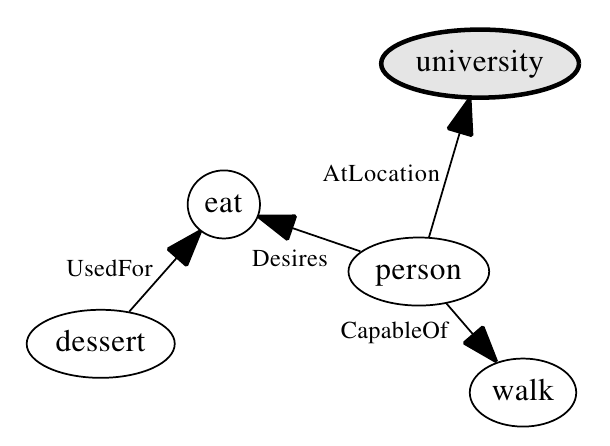}}}
	    \caption{Examples illustrating the types of commonsense mutation used in this study.}
	    \label{FigureMutation}
  	\end{figure*}

	\emph{Type I (concept attachment):} A new concept randomly picked from the set of concepts \emph{attachable} to the parent is attached through a new relation to one of existing concepts (Figure~\ref{FigureMutation} (a) and (b)).

	\emph{Type IIa (relation addition):} A new relation connecting two existing concepts in the parent is added, possibly connecting unconnected clusters within the same network (Figure~\ref{FigureMutation} (c) and (d)).

	\emph{Type IIb (relation deletion):} A randomly picked relation in the parent is deleted, possibly leaving unconnected clusters within the same network (Figure~\ref{FigureMutation} (e) and (f)).

	\emph{Type IIIa (concept addition):} A randomly picked new concept is added to the parent as a new cluster (Figure~\ref{FigureMutation} (g) and (h)).

	\emph{Type IIIb (concept deletion):} A randomly picked concept is deleted with all the relations it is involved in, possibly leaving unconnected clusters within the same network (Figure~\ref{FigureMutation} (i) and (j)).

	\emph{Type IV (concept replacement):} A concept in the parent, randomly picked from the set of those with at least one \emph{interchangeable} concept, is replaced with one (randomly picked) of its interchangeable concepts. Any relations left unsatisfied by the new concept are deleted (Figure~\ref{FigureMutation} (k) and (l)).

 \section{Results and Discussion}

	In this introductory study, we adopt values for crossover and mutation probabilities similar to earlier studies in graph-based EA \cite{Koza2003,Montes2004} (Table~\ref{TableParameters}). We use a crossover probability of $P_c = 0.85$, and a somewhat-above-average mutation rate of $P_m = 0.15$, accounting for the high tendency of mutation postulated in memetic literature\footnote{See Gil-White \cite{GilWhite2008} for a review and discussion of mutation in memetics.}. In our experiments, we subject a population of $Pop_{size} = 200$ individuals to tournament selection with tournament size $S_{size} = 8$ and winning probability $S_{prob} = 0.8$.

\begin{figure}[!t]
	    \centering
	    \subfigure[Given semantic network, 10 concepts, 11 relations (base domain)]{\parbox[t]{3.4in}{\centering\includegraphics[width=2.6in]{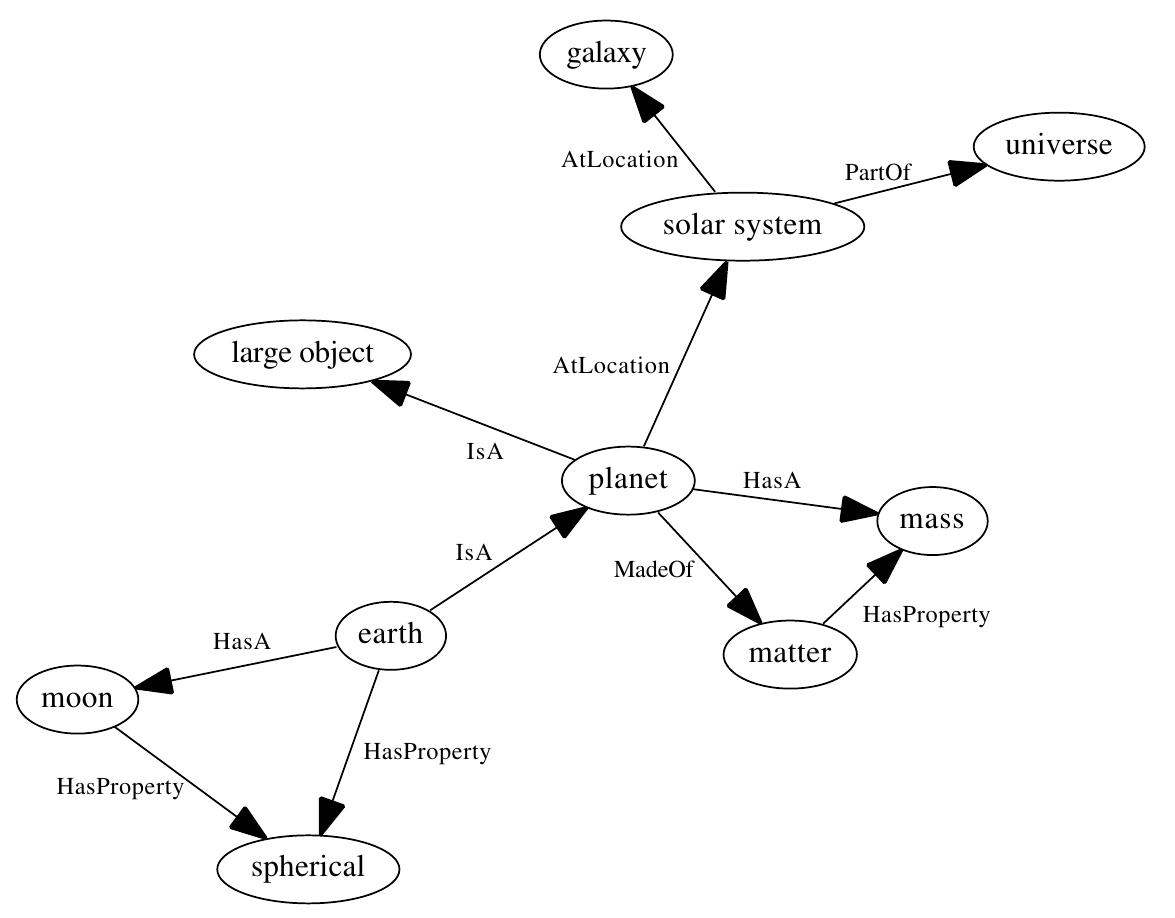}}}
	    \hfil
	    \subfigure[Evolved individual, 9 concepts, 9 relations (target domain)]{\parbox[t]{3.4in}{\centering\includegraphics[width=2.2in]{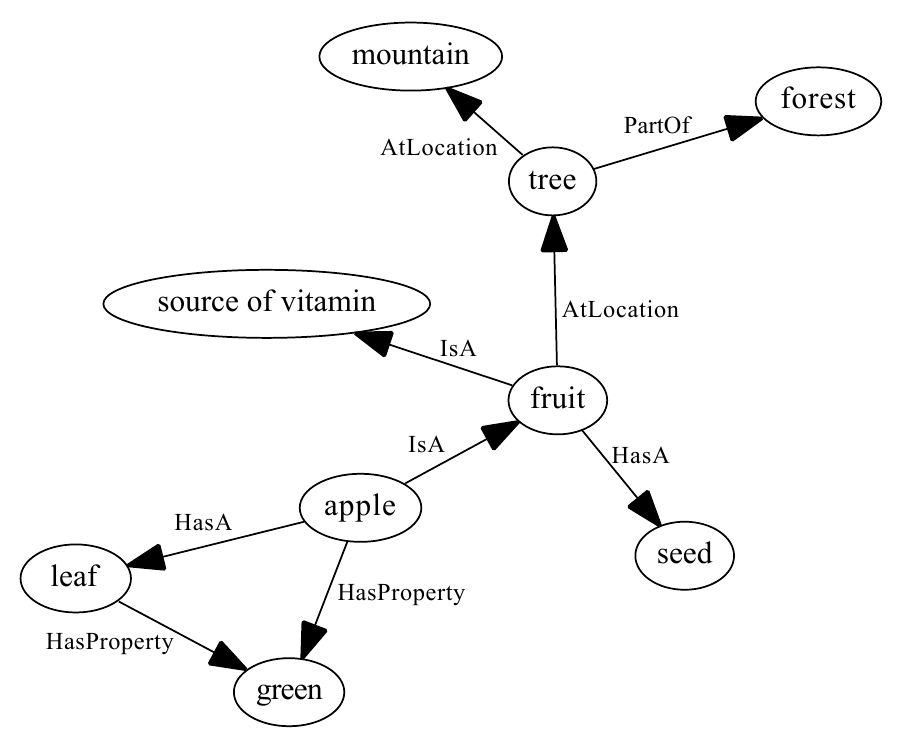}}}
	    \caption{Experiment 1: The evolved individual is encountered after 35 generations, with fitness value 2.8. Concepts and relations of the individual not involved in the analogy are not shown here for clarity.}
	    \label{FigureExperiment1}
  	\end{figure}

\begin{figure}[!t]
	    \centering
	    \subfigure[Given semantic network, 11 concepts, 11 relations (base domain)]{\parbox[t]{3.4in}{\centering\includegraphics[width=2.2in]{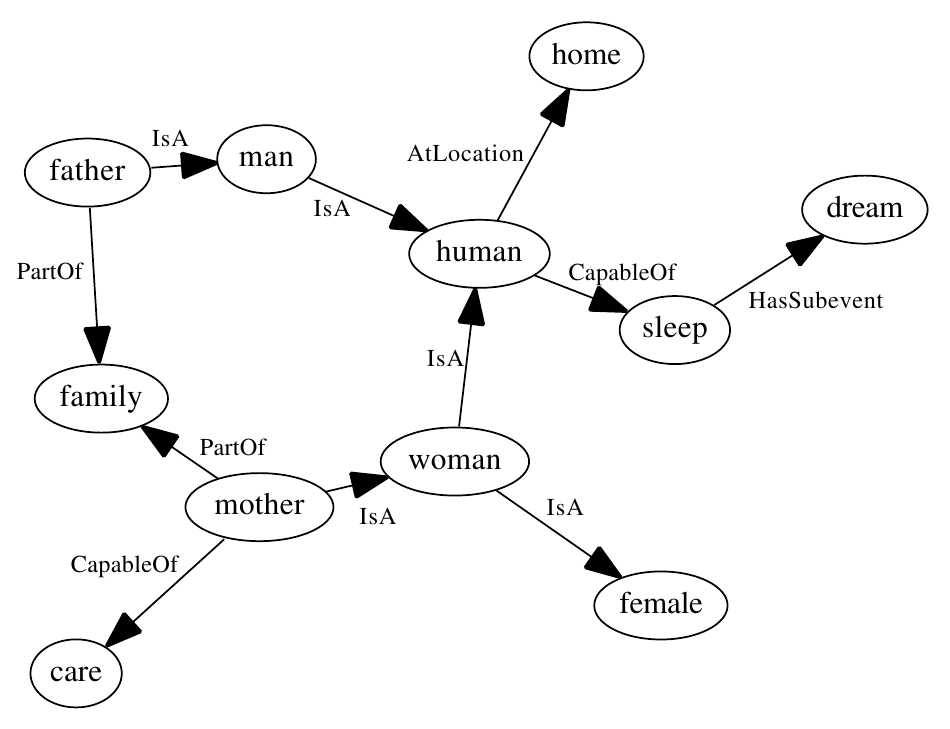}}}
	    \hfil
	    \subfigure[Evolved individual, 10 concepts, 9 relations (target domain)]{\parbox[t]{3.4in}{\centering\includegraphics[width=2.6in]{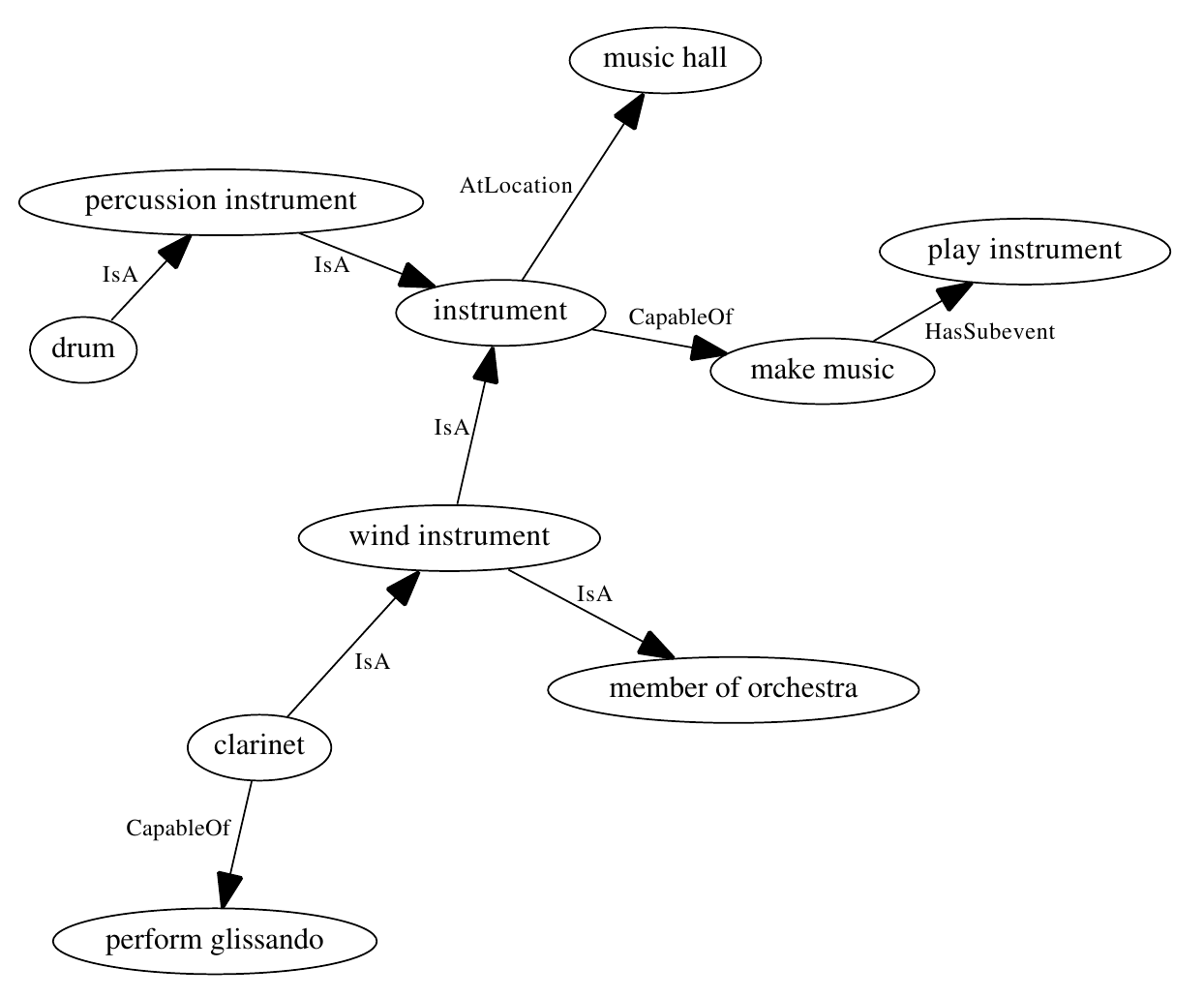}}}
	    \caption{Experiment 2: The evolved individual is encountered after 42 generations, with fitness value 2.7. Concepts and relations of the individual not involved in the analogy are not shown here for clarity.}
	    \label{FigureExperiment2}
  	\end{figure}

\begin{figure}[!t]
	    \centering
	    \subfigure[]{\includegraphics[width=2.6in]{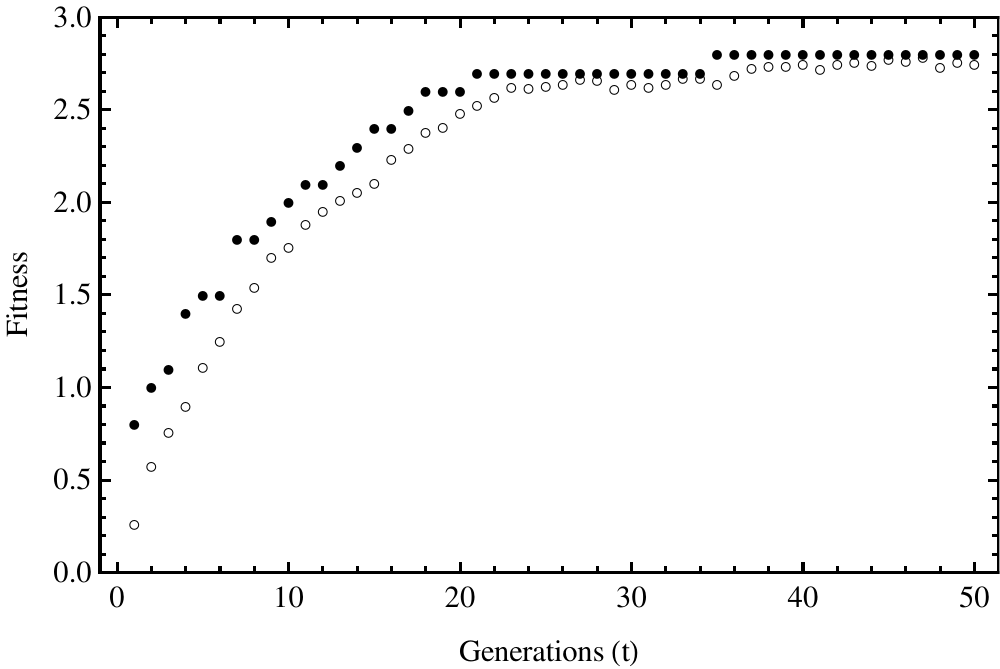}}
	    \subfigure[]{\includegraphics[width=2.6in]{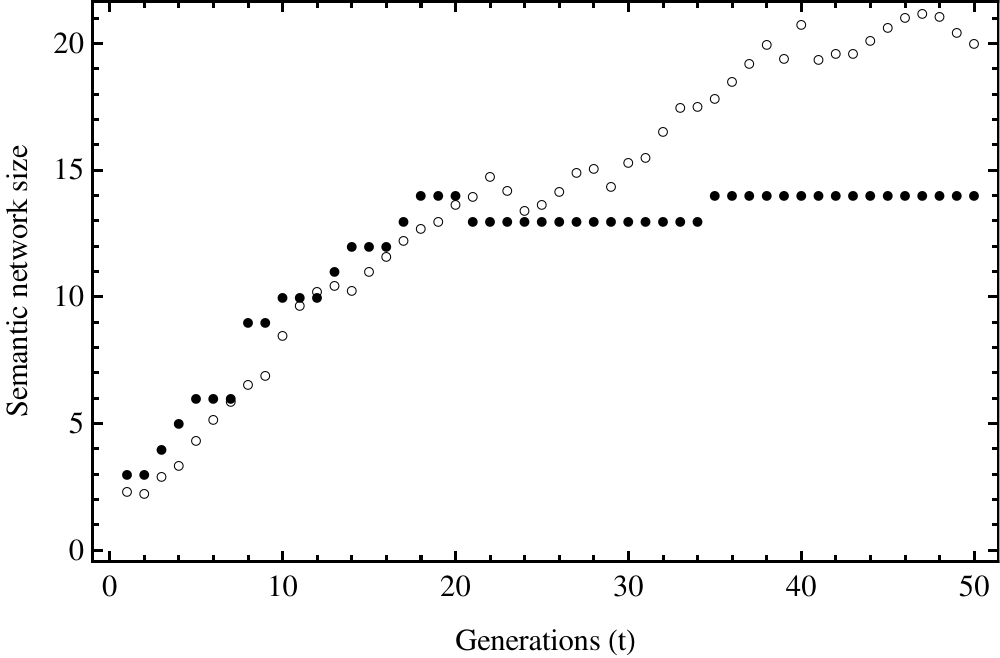}}
	    \caption{Evolution of (a) fitness and (b) semantic network size during the course of an experiment with parameters given in Table~\ref{TableParameters}. Filled circles represent the best individual in a generation, empty circles represent population average. Network size is taken to be the number of relations (edges).}
	    \label{FigurePlots}
  	\end{figure}

	Using this parameter set, we present the results from two runs of experiment: evolved analogies for a network describing some basic astronomical knowledge are shown in Figure~\ref{FigureExperiment1} and for a network of familial relations in Figure~\ref{FigureExperiment2}. We show in Figure~\ref{FigurePlots} (a) the progress of the best and average fitness in the population during the run that produced the results in Figure~\ref{FigureExperiment1}. The best and average size of semantic networks forming the individuals are shown in Figure~\ref{FigurePlots} (b). We observe that evolution asymptotically reaches a fitness plateau after about 40 generations. This coincides roughly with the point where the size of the best individual (13--14) becomes comparable with that of the given base semantic network (11, in Figure~\ref{FigureExperiment1}), after which improvements in the one-to-one analogy become sparser and less feasible. We also note that, between generations 21--34, the best network size actually gets smaller, demonstrating the possibility of improvement in network configuration without adding further nodes. Our experiments demonstrate that the proposed algorithm is capable of spontaneously creating collections of knowledge analogous to the one given in a base semantic network, with very good performance. In most cases, our implementation was able to reach extensive analogies within 50 generations and reasonable computational time.

	From an analogical reasoning viewpoint, the algorithm achieves the generation of diverse novel cases analogous to a given case. Compared with the Kilaza model of O'Donoghue \shortcite{ODonoghue2004} for finding novel analogous cases, which works by evaluating possible analogies to a given target case from a collection of candidate source domains that are assumed to be available, our approach is capable of open-ended and spontaneous creation of analogous cases from the ground up, replicating an essential mode of creative behavior observed in psychology \cite{Clement1988}.

	An important result is that, even if the use of commonsense knowledge in our algorithm was prompted by concerns that are practical in nature (i.e. restrictions on the meaningfulness and consistency of memetic variation by the introduced crossover and mutation operators), it eventually serves to tackle a very fundamental and long-standing problem in computational creativity: as put forth by Boden \shortcite{Boden2009}, ``no current AI system has access to the rich and subtly structured stock of concepts that any normal adult human being has built up over a lifetime'' and ``what's missing, as compared with the human mind, is the rich store of world knowledge (including cultural knowledge) that's often involved.'' We believe that the inherent commonsense reasoning element in our approach provides a means to address this criticism of lack of world knowledge in computational approaches to creativity.

 	\begin{table}[!t]
	    \caption{Parameters used during experiments}
	    \label{TableParameters}
	    \centering
	    \begin{tabular}{lcc}
	    \FL
	     & Parameter & Value
	    \ML
	    Evolution & Population size ($Pop_{size}$) & 200
	    \NN
	     & Crossover probability ($P_c$) & 0.85
	    \NN
	     & Mutation probability ($P_m$) & 0.15
	    \ML
	    Semantic & Max. initial concepts ($C_{max}$) & 5
	    \NN
	    networks & Min. relation score ($R_{min}$) & 2.0
	    \NN
	     & Timeout ($T$) & 10
	    \ML
	    Selection & Type & Tournament
	    \NN
	     & Tournament size ($S_{size}$) & 8
	    \NN
	     & Tournament win prob. ($S_{prob}$) & 0.8
	    \NN
	     & Elitism & Employed
	    \LL
	    \end{tabular}
  	\end{table}

\section{Conclusions and Future Work}

	We have presented a novel evolutionary algorithm that employs semantic networks as evolving individuals, paralleling the model of cultural evolution in the field of memetics. This algorithm, to our knowledge, is the first of its kind. The use of semantic networks provides a suitable basis for implementing variation and selection of memes as put forth by Dawkins \cite{Dawkins1989}. We have introduced preliminary versions of variation operators that work on this representation, utilizing knowledge from commonsense knowledge bases. We have also contributed a memetic fitness measure based on the structure mapping theory from psychology.

	Even if it is an intuitive fact that human culture and knowledge are evolving with time, existing models of culture, in their current state, are too minimalistic and weak in their descriptions of individual creativity and novelty; and conversely, theories modeling individual creativity lack consideration of cultural transmission and replication \cite{Gabora1997}. We believe that studies exploring creativity with evolutionary approaches have the potential for bridging this gap.

	In future work, an interesting possibility is to start the random semantic network generation procedure with several given concepts, allowing the discovery of cases formed around a particular set of seed concepts. The simple fitness function used in this introductory study can be extended to take graph-theoretical properties of semantic networks into account, such as the number of nodes or edges, shortest path length, or the clustering coefficient. The research would also benefit from exploring different types of mutation and crossover, and grounding the design of such operators on existing theories of cultural transmission and variation, discussed in sociological theories of knowledge.

	A direct and very interesting application of our approach would be to devise experiments with realistically formed fitness functions modeling selectionist theories of knowledge, which remain untested until this time. One such theory is the \emph{evolutionary epistemology} of Campbell \cite{Bickhard2003}, describing the development of human knowledge and creativity through selectionist principles such as blind variation and selective retention (BVSR).

\section{Acknowledgments}

	This work was supported by a JAE-Predoc fellowship from CSIC, and the research grants: 2009-SGR-1434 from the Generalitat de Catalunya, CSD2007-0022 from MICINN, and Next-CBR TIN2009-13692-C03-01 from MICINN.

\bibliographystyle{iccc}
\bibliography{BaydinEvolutionICCC2012}

\end{document}